\documentclass[final]{template}

\usepackage{times}
\usepackage{epsfig}
\usepackage{graphicx}
\usepackage{amsmath}
\usepackage{amssymb}
\usepackage{mathtools}
\usepackage{url}
\usepackage{color}
\usepackage{float}
\usepackage{array}
\usepackage{multirow}
\usepackage{arydshln}
\usepackage{longtable}

\usepackage{lipsum}
\usepackage{adjustbox}
\usepackage{enumitem}
\usepackage{fancyhdr}
\usepackage[pagebackref=true,breaklinks=true,colorlinks,bookmarks=false]{hyperref}

\providecommand{\keywords}[1]
{
  \small	
  \textbf{\textit{Keywords---}} #1
}

\begin{document}

\title{StressNAS: Affect State and Stress Detection Using Neural Architecture Search}

\author{Lam Huynh$^1$ \qquad
Tri Nguyen$^2$ \qquad
Thu Nguyen$^3$ \qquad
Susanna Pirttikangas$^2$ \qquad
Pekka Siirtola$^4$ \\
\small{$^1$Center for Machine Vision and Signal Analysis, University of Oulu} \qquad 
\small{$^2$Center for Ubiquitous Computing, University of Oulu} \\
\small{$^3$Economics and Business Administration, University of Oulu} \qquad 
\small{$^4$Biomimetics and Intelligent Systems Group, University of Oulu} }

\maketitle
\thispagestyle{fancy}
\lhead{\scriptsize To be published at ACM UbiComp-ISWC'21 Adjunct.}

\begin{abstract}

Smartwatches have rapidly evolved towards capabilities to accurately capture physiological signals. As an appealing application, stress detection attracts many studies due to its potential benefits to human health. It is propitious to investigate the applicability of deep neural networks (DNN) to enhance human decision-making through physiological signals. However, manually engineering DNN proves a tedious task especially in stress detection due to the complex nature of this phenomenon. To this end, we propose an optimized deep neural network training scheme using neural architecture search merely using wrist-worn data from WESAD. Experiments show that our approach outperforms traditional ML methods by $8.22\%$ and $6.02\%$ in the three-state and two-state classifiers, respectively, using the combination of WESAD wrist signals. Moreover, the proposed method can minimize the need for human-design DNN while improving performance by $4.39\%$ (three-state) and $8.99\%$ (binary).

\end{abstract}


\section{Introduction}
Long-term stress can have negative effects on both mental and physical human's health. This can even lead to economic cost, such as absenteeism, diminished productivity at work, accidents, etc.~\cite{halkos2010effect}. Thus, detecting stress can greatly contribute to improving human's health and adding value to the economy. 
Therefore, detecting stress in a person has been widely discussed, and using physiological changes in the human body is one of the approaches in stress detection. However, the topic of detecting other affective states has not been  seriously taken despite its contribution to human's emotion studies and commercial purposes. 

Gjoreski et al.~\cite{gjoreski2016continuous} is one of the first works that studied stress detection with a minimally intrusive approach. They used acceleration, blood volume pulse, heart rate data, galvanic skin response, and skin temperature recorded from a wrist-worn device to train a model for stress detection. Schmidt et al.~\cite{schmidt2018introducing} introduced a Multimodal Dataset for Wearable Stress and Affect Detection (WESAD). The data set contains physiological and motion data collected in a wrist-worn device and a chest-worn device. They trained five machine learning models for detecting different affective states  (baseline, stress and amusement): K-Nearest Neighbour (KNN), Linear Discriminant Analysis (LDA), Random Forest (RF), Decision Tree (DT), and AdaBoost Decision Tree (AB). The performance varies between the models and the types of classification problems. For the three-class classification problem (amusement vs. baseline vs. stress), the most well-performed approach when using wrist data is AB (75.21\%). For binary classification problem (stress vs. non-stress), the most achievable accuracy is 87.12\% obtained from RF using wrist data.

On the other hand, one might expect to improve affective detection performance by employing modern DNN architectures~\cite{simonyan2014very,he2016deep,iandola2016squeezenet,howard2017mobilenets,vaswani2017polosukhin}. However, despite the efforts~\cite{gjoreski2020machine,dziezyc2020can}, this remains a challenging task due to the complexity of human physiological signals. In this work, we introduced a novel framework, namely StressNAS, to optimize the deep neural network training using neural architecture search~\cite{barret2017neural}. For comparison, we also implement other DNN architectures, including a multilayer perceptron, a  fully convolutional network,  and a  residual-like  DNN. Moreover, wrist-worn devices are less intrusive to the human body than chest-worn ones and are more common among users (e.g., smartwatches,  wristbands). Potentially, affective state detection studies on data collected from wrist-worn devices can generate more user efficiency values. Therefore, we chose to evaluate our works in the three-state classifier (stress, baseline, and amusement) and binary classifier (stress and non-stress).

\begin{figure}[t!]
\begin{center}
  \includegraphics[width=0.9\linewidth]{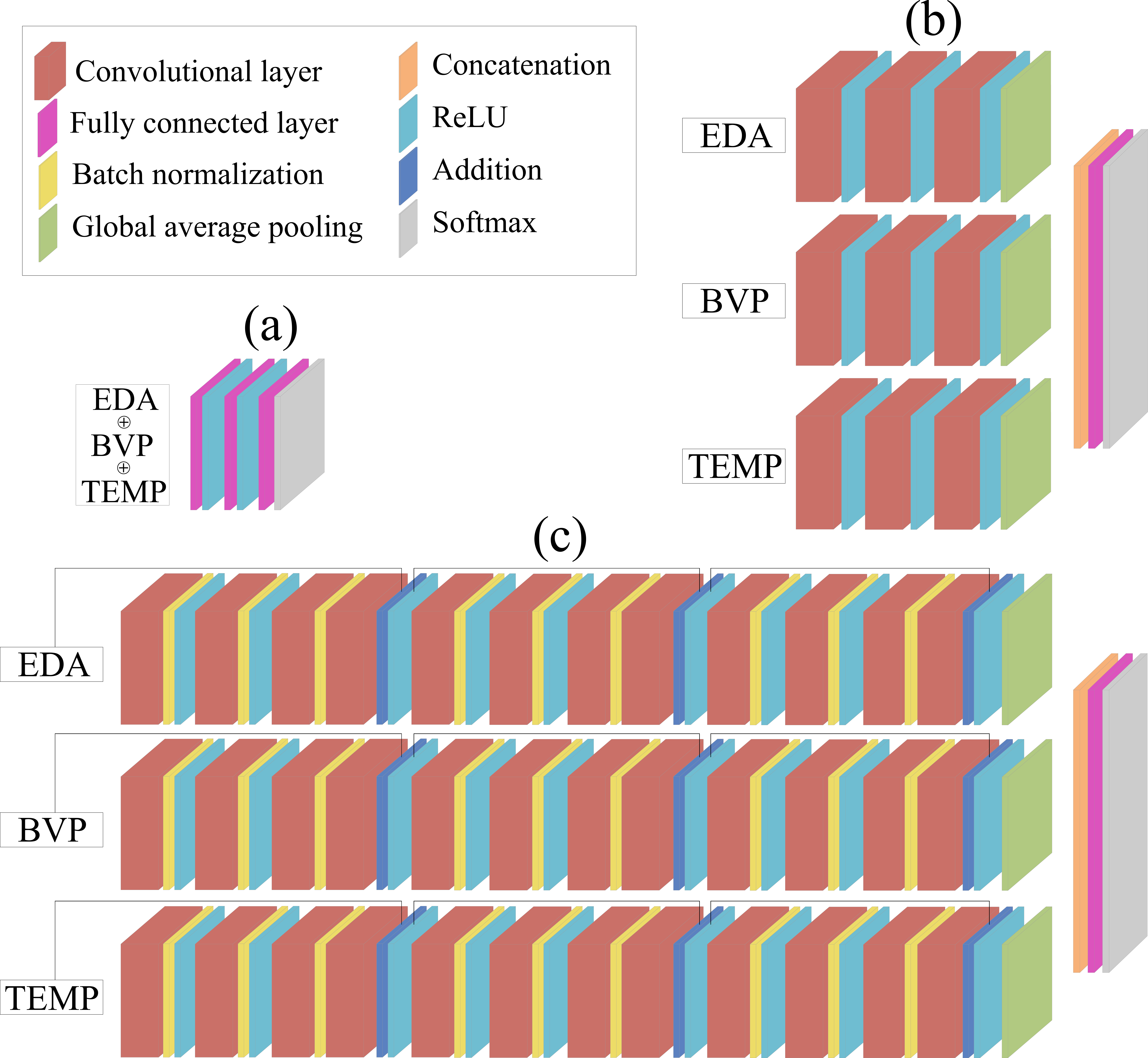}
\end{center}
  \caption{Manual design deep neural networks. 
  (a) Multilayer Perceptron, 
  (b) Fully Convolutional Network, 
  and (c) Residual-like Deep Neural Network.}
\label{fig:other_dnns}
\end{figure}

\section{Data}  \label{sec:data}

\noindent The Wearable Stress and Affect Detection (WESAD)~\cite{schmidt2018introducing} is a high-quality multi-modal dataset aiming for human affective detection. Participants were asked to follow standard protocols to calibrate their states (neural, stress, amusement, meditation) before their physiology and motion signals were captured. The data were collected from 17 subjects; each took part in a 2-hour section. Unfortunately, due to device malfunction, data from subject ID \#1 and \#12 were discarded.

\begin{figure*}[t!]
  \includegraphics[width=0.99\textwidth]{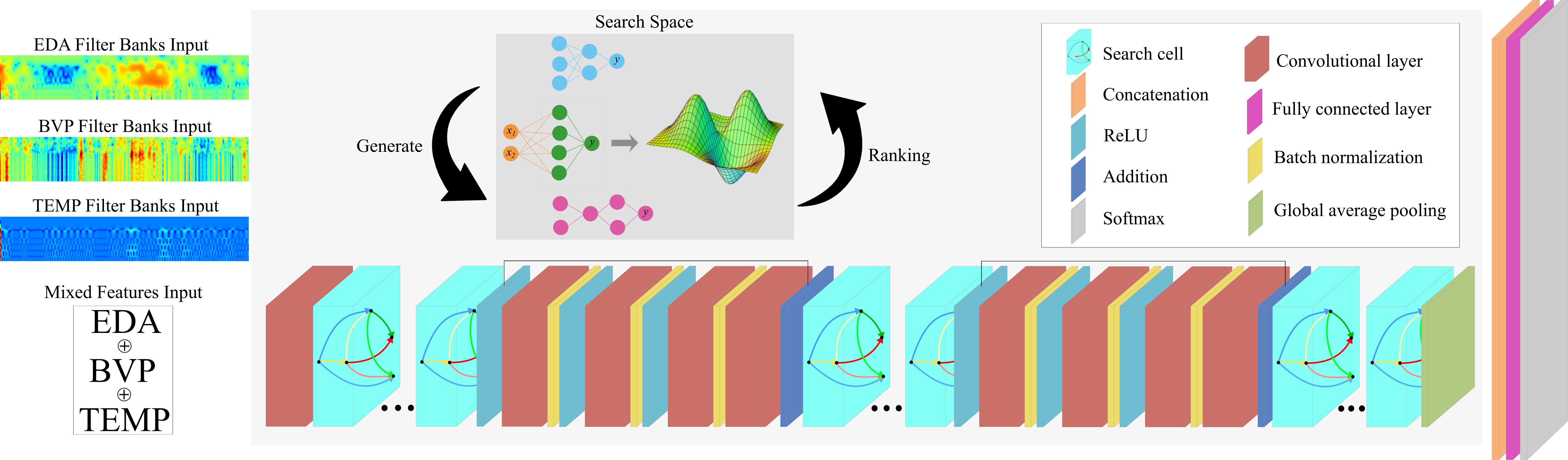}
  \caption{Overview architecture of StressNAS. The deep neural network (DNN) takes in EDA, BVP, TEMP filter banks, and mixed features. For each input, multiple DNNs are 1) randomly generated from the search space and 2) ranked based on their scores. Architectures with the highest-ranking are utilized for training. Output features of each modality are concatenated for final prediction.}
  \label{fig:stress_nas}
\end{figure*}

The data acquisition process utilized the chest-worn RespiBAN and wrist-worn Empatica E4 devices\footnote{www.empatica.com/research/e4/}. The RespiBAN provided: respiration (RESP), electrocardiogram (ECG), electrodermal activity (EDA), electromyogram (EMG), skin temperature (TEMP), and three-axis acceleration (ACC) at 700 Hz. On the other hand, the Empatica E4 measured blood volume pulse (BVP, 64 Hz), electrodermal activity (EDA, 4 Hz), body temperature (TEMP, 4 Hz), and three-axis acceleration (ACC, 32 Hz). The state conditions elicited from the protocol are referred to as the ground truth labels. As aforementioned, this work concentrates on the analysis of the WESAD wrist dataset.

\section{Methods}   \label{sec:method}
This work uses a set of manual design DNNs with the WESAD dataset for preprocessing, training, and evaluating. 
Furthermore, we propose StressNAS, an automatic neural architecture search for affective state prediction.  

\subsection{Data processing}
As mentioned earlier, we mainly utilized the data from the Empatica E4 for experimenting with the three-state (baseline, stress, and amusement) and the two-state (stress and non-stress) classification problems. 
We argue that training models with wrist data alone can be more challenging due to large variations in sampling rates of input modalities. 
Besides, we also transform the time-series data to filter banks for training DNNs.

The data were sampled using the sliding window technique. All experiments were conducted using a window length of 60 seconds with a 0.25-second shifting. As a result, approximately 132600 data samples were created in total.

A filter bank is a quadratic form of signal in the joint time-frequency domain that is a popular representation for training DNNs in speech processing. To obtain the filter banks, one can: 1) pass the signal through a pre-emphasis filter, 2) acquire overlapping frames from the filtered signal and then apply a windowing function (e.g. Hamming window), 3) take the Short-Time Fourier-Transform to get the power spectrum, 4) apply triangular filters and mean normalization to calculate the filter banks. 

\begin{table*}[t!]
\caption{\label{tab:eval_three_state}Results of different models and sensor combinations for classifying neural v. stress v. amusement states. Abbreviations: KNN = k-nearest neighbour, AB = AdaBoost DT, DT = Decision Tree, LDA = Linear discriminant analysis, RF = Random Forest, MLP = Multilayer Perceptron, FCN = Fully Convolution Network, ResNet = ResNet-like DNN, and StressNAS = our proposal.} %
\centering
\small
\adjustbox{max width=\textwidth}{\begin{tabular}{|l|c|c|c|c|c|c|c|c|c|}
\hline
\multirow{2}{*}{Sensor Combinations}  &  \multicolumn{5}{c|}{Schmidt et al.~\cite{schmidt2018introducing}}                             & \multicolumn{3}{c|}{Manual Design DNNs} & \multirow{2}{*}{StressNAS}   \\ \cline{2-9} 
                                       &  AB & DT & RF & KNN & LDA & MLP & FCN & ResNet &  \\ \hline 
                                       
ACC+EDA+BVP+TEMP & 75.21 $\pm$ 0.77 & 53.98 $\pm$ 1.79 & 74.85 $\pm$ 0.20 & 45.54 & 70.74  & 78.11 & 79.04 & 79.48 & \textbf{83.43}  \\ \hline

EDA+BVP+TEMP     & 73.62 $\pm$ 0.55 & 63.34 $\pm$ 1.00 & 76.17 $\pm$ 0.42 & 58.54 & 68.85 & 73.60 & 74.12 & 73.93 & \textbf{81.78} \\ \hline

ACC & \textbf{57.07 $\pm$ 0.57} & 53.71 $\pm$ 0.91 & 56.40 $\pm$ 0.16 & 45.54 & 47.73 & 52.16 & 46.53 & 45.25 & 55.81 \\  %

EDA & 59.42 $\pm$ 0.27 &  54.36 $\pm$ 0.27 & 56.57 $\pm$ 0.05 & 54.98 & 62.32 & 57.36 & 62.14 & 63.50 & \textbf{66.89} \\ %

BVP & 64.46 $\pm$ 0.21 & 57.57 $\pm$ 0.22 & 64.09 $\pm$ 0.12 & 59.44 & 70.17 & 62.43 & 65.42 & 68.11 & \textbf{71.24} \\  %

TEMP & 49.39 $\pm$ 0.23 & 47.42 $\pm$ 0.36 & 48.67 $\pm$ 0.21 & 44.32  & 58.96  & 55.14 & 56.32 & 61.35 & \textbf{62.15} \\ \hline %

\end{tabular}}
\end{table*}

\begin{table*}[t!]
\caption{\label{tab:eval_two_state}Results of different models and sensor combinations for classifying stress v. non-stress. Abbreviations: KNN = k-nearest neighbour, AB = AdaBoost DT, DT = Decision Tree, LDA = Linear discriminant analysis, RF = Random Forest, MLP = Multilayer Perceptron, FCN = Fully Convolution Network, ResNet = ResNet-like DNN, and StressNAS = our proposal.} %
\centering
\small
\adjustbox{max width=\textwidth}{\begin{tabular}{|l|c|c|c|c|c|c|c|c|c|}
\hline
\multirow{2}{*}{Sensor Combinations}  &  \multicolumn{5}{c|}{Schmidt et al.~\cite{schmidt2018introducing}}                                & \multicolumn{3}{c|}{Manual Design DNNs} & \multirow{2}{*}{StressNAS}   \\ \cline{2-9} 
                                       &  AB & DT & RF & KNN & LDA & MLP & FCN & ResNet &  \\ \hline 

ACC+EDA+BVP+TEMP & 83.98 $\pm$ 0.75 & 82.19 $\pm$ 0.44 & 87.12 $\pm$ 0.24 & 63.80  & 86.88 & 83.19 & 84.15 & 83.14 & \textbf{93.14}  \\ \hline

EDA+BVP+TEMP     & 88.05 $\pm$ 0.18 & 84.88 $\pm$ 0.11 & 88.33 $\pm$ 0.25 & 81.96  & 86.46 & 82.12 & 82.31 & 82.77 & \textbf{92.87} \\ \hline

ACC              & 71.69 $\pm$ 0.45  & 64.08 $\pm$ 0.49 & 69.96 $\pm$ 0.42 & 63.80 & 60.02 & 65.15 & 67.88 & 66.85 & \textbf{72.15} \\ 

EDA              & \textbf{79.71 $\pm$ 0.43}  & 76.21 $\pm$ 0.27 & 76.29 $\pm$ 0.14 & 73.13 & 78.08 & 62.78 & 69.13 & 67.81 & 79.24 \\ 

BVP              & 84.10 $\pm$ 0.13  & 81.39 $\pm$ 0.15 & 84.18 $\pm$ 0.11 & 82.06 & \textbf{85.83} & 69.97 & 72.15 & 69.15 & 81.16 \\ 

TEMP             & 67.11 $\pm$ 0.34  & 68.22 $\pm$ 0.19 & 67.82 $\pm$ 0.11 & 64.46 & 69.24 & 55.17 & 68.12 & 62.54 & \textbf{71.46} \\ \hline

\end{tabular}} 
\end{table*}

\subsection{Manual design deep neural networks}

Figure~\ref{fig:other_dnns} shows the architecture of a multilayer perceptron (MLP), a fully convolutional network (FCN)~\cite{long2015fully}, and a residual-like DNN (ResNet)~\cite{he2016deep} used in our experiments. The MLP consists of three fully connected (FC) layers follow by nonlinear activation functions (rectified linear unit (ReLU) after layer 1-2 and Softmax for the last layer). 
The FCN contains several branches for separate modalities. Each branch has three convolutional layers (CONV), follow by ReLU with global average pooling (GAP) at the last layer. These features are then concatenated before feeding to an FC and Softmax for final prediction.
The ResNet inherits a similar design principle of FCN while expanding the network using the residual blocks (Res-block). The Res-block contains four CONV-layers with batch normalization, ReLU, and residual connection. Each ResNet branch has three Res-block follow by a GAP before final concatenation and prediction.

\subsection{StressNAS}

The proposed network includes multiple DNNs that take in EDA, BVP, TEMP filter banks, and mixed features as inputs. Instead of manually constructing the DNN, we first randomly generated a set of network candidates (DNNs) for each input modality from the search space following ~\cite{dong2019bench} procedure. The core component of our candidates is the search cell. It is a directed acyclic graph that contains densely connected edges (feature transform operations, e.g., convolution, pooling, skip-connection) between its nodes (computed tensors). 

For each modality, we randomly search from 10000 architectures. We then rank these architectures based on their scores that is calculated as the covariance matrices of the gradient with respect to the input data at the initial time~\cite{mellor2020neural}. The best ten architectures with the highest-ranking are utilized for training. Output features of each modality are concatenated for final prediction. In total, the searching, ranking and training time of our proposed method on one Telsa-V100 is $\sim 50$ hours. Figure~\ref{fig:stress_nas} shows the overview architecture of our proposed approach.

\section{Experiments and Results}
The experiments consist of the evaluation of manual design DNNs and StressNAS models in three-state and two-state classifiers. Besides, those DNNs' results are compared with traditional models by Schmidt et al.~\cite{schmidt2018introducing}.

\subsection{Experiment and Evaluation metric}

The experiments are about the implementation of different DNN approaches. In particular, we implement and build MLP, FCN, ResNet, and especially our proposal StressNAS for a three-state classifier (stress, baseline, and amusement) and binary classifier (stress and non-stress). The binary classifier considers the stress records and non-stress records as the combination of baseline and amusement records. For evaluation, we separate our training and testing data using leave-one-out cross-validation on 15 subjects from the WESAD dataset to build user-invariant models. Particularly, each person, by turn, becomes the test set, and the other 14 samples' data are the training set. Hence, the balanced accuracy in the experiments is an average accuracy of 15 prediction results.

\subsection{Results}

The results of the experiments are described in Table~\ref{tab:eval_three_state} and Table~\ref{tab:eval_two_state}. In detail, the first column indicates the combination of sensors in the two first rows and the individual sensors in the remaining rows, while traditional approaches by Schmidt et al.~\cite{schmidt2018introducing} are illustrated in five first columns, and the other columns are DNNs' result. Remarkably, our proposal results with StressNAS are in the last column.

\begin{table}[t!]
\caption{\label{tab:subject_id} Best detection accuracies for each study subject for three states classification task.}
\centering
\small
\begin{tabular}{|c|c|}
\hline
\textbf{Study subject} & \textbf{Accuracy$\uparrow$} \\ \hline

2 & 81.07 \\ \hline
3 & 72.55 \\ \hline
4 & 82.82 \\ \hline
5 & 79.84 \\ \hline
6 & 89.50 \\ \hline
7 & 81.52 \\ \hline
8 & 81.84 \\ \hline
9 & 88.43 \\ \hline
10 & 74.55 \\ \hline
11 & 85.61 \\ \hline
13 & 75.55 \\ \hline
14 & 87.11 \\ \hline
15 & 91.40 \\ \hline
16 & 95.55 \\ \hline
17 & 84.20 \\ \hline

\end{tabular} 
\end{table}

Table~\ref{tab:eval_three_state} presents the prediction results of four DNNs models in a three-state classifier with stress, baseline, and amusement. When comparing manual DNNs with traditional machine learning methods, ResNet performs better in three sensor combinations (ACC + EDA + BVP + TEMP, EDA, and TEMP), while MLP and FCN tend to have better performance with ACC + EDA + BVP + TEMP setting. Automatic DNN StressNAS provides better prediction than traditional machine learning methods in most data combinations (except ACC data setting). Among different sensor combinations, both manual DNNs and StressNAS gain the highest accuracy with the combination of ACC, EDA, BVP, and TEMP. Specifically, all manual DNNs (MLP, FCN, and ResNet) and StressNAS obtain higher accuracy than  traditional machine learning methods by 3-4\% and 8\%, respectively, in the combination of ACC, EDA, BVP, and TEMP. Moreover, Table~\ref{tab:subject_id} presents the best detection accuracies for each study subject for three states classification task. In general, the detection figures are evenly distributed among our subjects with a slight bias on subjects 3, 10, 13, and 16.

Table~\ref{tab:eval_two_state} shows the experimental results of the binary classifier (stress and non-stress detection). As is the case for the three-state classifier, both manual DNNs and StressNAS perform best with the sensor combination of ACC, EDA, BVP, and TEMP. However, interestingly, different from the three-class classifier, manual DNNs' results are similar or lower than traditional models' results, especially LDA and AB's results. In contrast, the proposed method StressNAS provides a substantial improvement for almost all settings (except EDA and BVP signals). Notably, with the data combination of ACC, EDA, BVP, and TEMP, StressNAS achieves better prediction than traditional machine learning methods by 6\%.

\section{Conclusion}    \label{sec:conclu}
With the swift development of technologies, sensor devices have become more popular and wide uses in health care. Stress detection is a crucial topic supporting and improving human health by monitoring and analyzing body signals' changes. Due to the interest in stress detection through physiological signals, many studies have concentrated on different activities, including constructing datasets, applying signal processing, and utilizing machine learning models. Despite existing studies related to stress detection, applying deep learning approaches still lacks consideration from the community. Therefore, the paper details using deep learning approaches to detect stress state through physiological signals. In detail, the paper utilizes a set of deep neural approaches (MLP, FCN, and ResNet) and proposes StressNAS, an automatically detected neural network architecture, to analyze affective states from WESAD's wrist dataset. The evaluation is based on leave-one-out cross-validation to gain the result for comparison among different approaches in a three-state classifier (stress, baseline, and amusement) and binary classifier (stress and non-stress). Manual design DNNs are somewhat borderline with classical machine learning methods (CLASS), which suggests that carefully designing network architecture is a tedious task. On the other hand, DNN generated by neural architecture search (StressNAS) yields enhanced performance across all sensor combinations. Also, StressNAS tends to achieve better accuracy than CLASS in both the classifiers.

\section*{Acknowledgement}

\noindent This work is supported by the Academy of Finland 6Genesis Flagship (grant 318927), the Vision-based 3D perception for mixed reality applications project and the TrustedMaaS project by the Infotech institute of the University of Oulu.

\bibliographystyle{ieee_fullname}
\bibliography{ms}

\begin{thebibliography}{10}\itemsep=-1pt

\bibitem{barret2017neural}
Zoph Barret and V~Le Quoc.
\newblock Neural architecture search with reinforcement learning.
\newblock In {\em International conference on learning representatoins}, 2017.

\bibitem{dong2019bench}
Xuanyi Dong and Yi Yang.
\newblock Nas-bench-201: Extending the scope of reproducible neural
  architecture search.
\newblock In {\em International Conference on Learning Representations}, 2019.

\bibitem{dziezyc2020can}
Maciej Dzie{\.z}yc, Martin Gjoreski, Przemys{\l}aw Kazienko, Stanis{\l}aw
  Saganowski, and Matja{\v{z}} Gams.
\newblock Can we ditch feature engineering? end-to-end deep learning for affect
  recognition from physiological sensor data.
\newblock {\em Sensors}, 20(22):6535, 2020.

\bibitem{gjoreski2020machine}
Martin Gjoreski, Matja~{\v{Z}} Gams, Mitja Lu{\v{s}}trek, Pelin Genc, Jens-U
  Garbas, and Teena Hassan.
\newblock Machine learning and end-to-end deep learning for monitoring driver
  distractions from physiological and visual signals.
\newblock {\em IEEE Access}, 8:70590--70603, 2020.

\bibitem{gjoreski2016continuous}
Martin Gjoreski, Hristijan Gjoreski, Mitja Lu{\v{s}}trek, and Matja{\v{z}}
  Gams.
\newblock Continuous stress detection using a wrist device: in laboratory and
  real life.
\newblock In {\em proceedings of the 2016 ACM international joint conference on
  pervasive and ubiquitous computing: Adjunct}, pages 1185--1193, 2016.

\bibitem{halkos2010effect}
George Halkos and Dimitrios Bousinakis.
\newblock The effect of stress and satisfaction on productivity.
\newblock {\em International Journal of Productivity and Performance
  Management}, 2010.

\bibitem{he2016deep}
Kaiming He, Xiangyu Zhang, Shaoqing Ren, and Jian Sun.
\newblock Deep residual learning for image recognition.
\newblock In {\em Proceedings of the IEEE conference on computer vision and
  pattern recognition}, pages 770--778, 2016.

\bibitem{howard2017mobilenets}
Andrew~G Howard, Menglong Zhu, Bo Chen, Dmitry Kalenichenko, Weijun Wang,
  Tobias Weyand, Marco Andreetto, and Hartwig Adam.
\newblock Mobilenets: Efficient convolutional neural networks for mobile vision
  applications.
\newblock {\em arXiv preprint arXiv:1704.04861}, 2017.

\bibitem{iandola2016squeezenet}
Forrest~N Iandola, Song Han, Matthew~W Moskewicz, Khalid Ashraf, William~J
  Dally, and Kurt Keutzer.
\newblock Squeezenet: Alexnet-level accuracy with 50x fewer parameters and< 0.5
  mb model size.
\newblock {\em arXiv preprint arXiv:1602.07360}, 2016.

\bibitem{long2015fully}
Jonathan Long, Evan Shelhamer, and Trevor Darrell.
\newblock Fully convolutional networks for semantic segmentation.
\newblock In {\em Proceedings of the IEEE conference on computer vision and
  pattern recognition}, pages 3431--3440, 2015.

\bibitem{mellor2020neural}
Joseph Mellor, Jack Turner, Amos Storkey, and Elliot~J. Crowley.
\newblock Neural architecture search without training, 2020.

\bibitem{schmidt2018introducing}
Philip Schmidt, Attila Reiss, Robert Duerichen, Claus Marberger, and Kristof
  Van~Laerhoven.
\newblock Introducing wesad, a multimodal dataset for wearable stress and
  affect detection.
\newblock In {\em Proceedings of the 20th ACM International Conference on
  Multimodal Interaction}, pages 400--408, 2018.

\bibitem{simonyan2014very}
Karen Simonyan and Andrew Zisserman.
\newblock Very deep convolutional networks for large-scale image recognition.
\newblock {\em arXiv preprint arXiv:1409.1556}, 2014.

\bibitem{vaswani2017polosukhin}
A Vaswani, N Shazeer, N Parmar, J Uszkoreit, L Jones, and AN Gomez.
\newblock \& polosukhin, i.(2017). attention is all you need.
\newblock {\em Advances in neural information processing systems}, pages
  5998--6008, 2017.

\end{thebibliography}

\end{document}